\documentclass[10pt,twocolumn,letterpaper]{article}

\usepackage{iccv}              

\definecolor{defaultcolor}{gray}{.9}
\newcommand{\default}[1]{\cellcolor{defaultcolor}{#1}}

\renewcommand{\paragraph}[1]{\noindent\textbf{#1}}

\definecolor{iccvblue}{rgb}{0.21,0.49,0.74}
\usepackage[pagebackref,breaklinks,colorlinks,allcolors=iccvblue]{hyperref}

\usepackage{microtype}
\usepackage{graphicx}
\usepackage{subcaption}
\usepackage{booktabs} 
\usepackage{multirow}
\usepackage{multicol}
\usepackage[table,xcdraw]{xcolor}
\usepackage{arydshln}

\usepackage[capitalize,noabbrev]{cleveref}

\usepackage{pifont}
\newcommand{\cmark}{\ding{51}}%
\newcommand{\xmark}{\ding{55}}%
\newcommand{\argmin}{\arg\min}
\newcommand{\ours}{CODA }
\newcommand{\tablestyle}[2]{\setlength{\tabcolsep}{#1}\renewcommand{\arraystretch}{#2}\centering\footnotesize}
\newlength\savewidth\newcommand\shline{\noalign{\global\savewidth\arrayrulewidth
  \global\arrayrulewidth 1pt}\hline\noalign{\global\arrayrulewidth\savewidth}}

\usepackage{listings}
\newcommand\codeurl[1]{{{\color{blue}{\url{#1}}}}}
\usepackage{xcolor}
\definecolor{commentgreen}{rgb}{0.4167, 0.60196, 0.3333}

\usepackage{algorithm}
\usepackage{algpseudocode}
\usepackage[accsupp]{axessibility}  

\def\paperID{164} 
\def\confName{ICCV}
\def\confYear{2025}

\title{CODA: Repurposing Continuous VAEs for Discrete Tokenization}

\author{\quad Zeyu Liu$^1$\thanks{Equal contribution.}\quad\quad\;Zanlin Ni$^{1*}$\quad\quad\;Yeguo Hua$^{1}$\quad\quad\;Xin Deng$^{2}$\\
Xiao Ma$^{3}$\quad\quad\;Cheng Zhong$^{3}$\quad\quad\;Gao Huang$^{1}\thanks{Corresponding Author.}$ \\ 
{\small $^1$ Tsinghua University}\quad\quad
{\small $^2$ Renmin University}\quad\quad
{\small $^3$ Lenovo Research, AI Lab}\\
{\footnotesize\codeurl{{https://lzy-tony.github.io/coda}}}\vspace{-2mm}}

\begin{document}
\maketitle
\begin{abstract}
Discrete visual tokenizers transform images into a sequence of tokens, enabling token-based visual generation akin to language models. However, this process is inherently challenging, as it requires both \emph{compressing} visual signals into a compact representation and \emph{discretizing} them into a fixed set of codes.
Traditional discrete tokenizers typically learn the two tasks jointly, often leading to unstable training, low codebook utilization, and limited reconstruction quality. In this paper, we introduce \textbf{\ours}(\textbf{CO}ntinuous-to-\textbf{D}iscrete \textbf{A}daptation), a framework that decouples compression and discretization.
Instead of training discrete tokenizers from scratch, \ours adapts off-the-shelf continuous VAEs---already optimized for perceptual compression---into discrete tokenizers via a carefully designed discretization process.
By primarily focusing on discretization, \ours ensures stable and efficient training while retaining the strong visual fidelity of continuous VAEs.
Empirically, with $\mathbf{6 \times}$ less training budget than standard VQGAN, our approach achieves a remarkable codebook utilization of \textbf{100\%} and notable reconstruction FID (rFID) of $\mathbf{0.43}$ and $\mathbf{1.34}$ for $8 \times$ and $16 \times$ compression on ImageNet 256$\times$ 256 benchmark.

\end{abstract}
    
\section{Introduction}
\label{sec:intro}

\begin{figure}[t]
    \includegraphics[width=\linewidth]{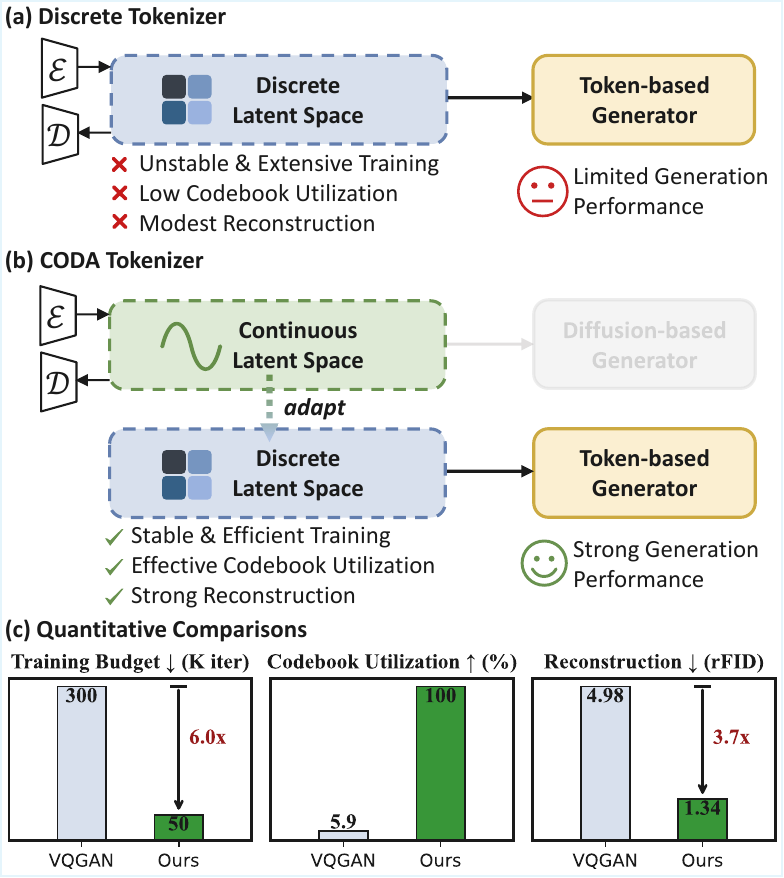}
    \caption{(a) \textbf{Conventional discrete VQ tokenizers} learn to \emph{compress} and \emph{discretize} inherently continuous visual signals into codes simultaneously. This lead to multiple challenges in training and the corresponding unsatisfactory latent space poses a bottleneck that limit the performance of discrete token-based generation models. (b) \textbf{Our proposed \ours tokenizer} leverages continuous VAEs for compression, directly discretizing the latent space. (c) \textbf{Quantitative comparisons} between VQGAN~\cite{esser2021taming} and our proposed \ours tokenizer.}
    \label{fig:main_idea}
\end{figure}

The field of AI-generated content (AIGC) has witnessed significant progress in recent years. In natural language processing, text generation has been mainly unified by the paradigm of next discrete token prediction~\cite{vaswani2017attention, raffel2020exploring, brown2020language, achiam2023gpt}.
This contrasts with developments in computer vision, where the debate between \textit{continuous} and \textit{discrete} generation paradigms has yet to reach a conclusion. 
While continuous diffusion models dominate the field~\cite{rombach2022high, podellsdxl, peebles2023scalable, flux}, discrete token-based approaches~\cite{chang2022maskgit, chang2023muse, ni2024revisiting, sun2024autoregressive, tian2024visual} are increasingly gaining interest due to their computational efficiency~\cite{tian2024visual, nienat, Ni2024AdaNAT} and promising potential for unifying tasks across language modeling and multimodal understanding~\cite{xie2024showo, wang2024emu3, wu2024vila, wu2024janus}.

One defining characteristic of discrete methods is the need for a discrete tokenizer, which converts visual signals into a discrete format akin to language tokens. This process, however, is non-trivial, as it requires simultaneously \emph{compressing} visual signals and \emph{discretizing} them into a set of codes. In contrast, continuous VAEs focus solely on mapping visual signals into a compressed continuous latent space, avoiding this added complexity.
Previous research has identified various problems in standard VQ-based discrete tokenizers, including unstable training~\cite{yu2021vector, baykal2024edvae, zhustabilize}, low codebook utilization~\cite{fifty2024restructuring, zhu2024scaling, huh2023straightening}, and limited performance even under extensive training~\cite{esser2021taming, webermaskbit}.
For example, while VAE used in state-of-the-art diffusion models~\cite{flux} achieves $0.17$ reconstruction FID (rFID) on ImageNet~\cite{deng2009imagenet}, the standard VQGAN~\cite{esser2021taming} tokenizer achieves only $4.98$ rFID.
In response to these challenges, recent research has explored many advanced techniques for discrete tokenization, such as embedding-free tokenization~\cite{luo2024open, mentzerfinite, yulanguage, webermaskbit}, rotation trick~\cite{fifty2024restructuring}, and 1D tokenization~\cite{ge2023planting, yu2024image}.

Despite these recent advances, discrete tokenizers are still required to learn compression and discretization concurrently and are often regarded as mutually exclusive from continuous VAEs.
In this paper, we challenge this dichotomy and propose a novel \textbf{\ours}(\textbf{CO}ntinuous-to-\textbf{D}iscrete \textbf{A}daptation) framework that decouples compression and discretization.
Instead of training a discrete tokenizer to handle compression and discretization in tandem, we demonstrate that off-the-shelf continuous VAEs, which are already highly optimized for perceptual compression, can be directly adapted into discrete tokenizers through a carefully designed discretization process (see~\cref{tab:design_analysis} and~\cref{fig:tokenizer_structure}).
In this way, \ours retains the strong visual fidelity of continuous VAEs while enabling discrete representation through a secondary transformation.
Moreover, by focusing primarily on discretization, our method ensures more stable and efficient training, leading to significantly higher codebook utilization and improved reconstruction performance.

Building upon this conceptual framework, we craft a set of \ours tokenizers, instantiated through systematic adaptations on the $16 \times$ compression MAR~\cite{li2024autoregressive} and $8 \times$ compression FLUX~\cite{flux} VAEs. Quantitative evaluations demonstrate that \ours achieves competitive performance of $1.34$ and $0.43$ rFID on the ImageNet dataset, respectively, while maintaining full codebook utilization. Moreover, our proposed disentangling leads to high computational efficiency in training, enabling a reduction of $6 \times$ in training compute compared to representative prior VQGAN training settings. Empirical results also demonstrate that, when integrated with the \ours tokenizer, representative discrete token-based generation methods, \ie MaskGIT~\cite{chang2022maskgit} can be significantly improved to achieve overall competitive performance against main stream continuous and discrete generation paradigms.

Our main contributions can be summarized as follows:

\begin{enumerate}
    \item We challenge the conventional dichotomy of continuous VAEs and discrete tokenizers, and present a novel prospective to designing discrete tokenizers by leveraging continuous VAEs for compression while learning solely effective discretization.
    \item Motivated by this approach, we craft a series of \ours tokenizers, reaching competitive reconstruction quality while requiring only a fraction of training time compared to standard settings.
    \item Combining with MaskGIT, we demonstrate that our \ours tokenizer unlocks enhanced and competitive performance in discrete token-based image synthesis.
\end{enumerate}

\section{Related Work}
\label{sec:related_work}

\subsection{Continuous and Discrete Image Synthesis}

\noindent\textbf{Continuous}
approaches to image synthesis, i.e. latent diffusion models~\cite{rombach2022high, podellsdxl, peebles2023scalable, flux} operate on a latent space sampled from a Variational Autoencoder (VAE)~\cite{kingma2013auto}. Images are first mapped from pixel space into this compressed representation and then subjected to iterative denoising via diffusion models. At each step, a trained neural network refines the latent by removing noise and rebuilding semantic structure and fine details. Once denoising is complete, the latents are decoded back into pixel space to generate the final image. This approach has been proven highly successful, enabling the generation of high-quality, photorealistic images.

\noindent\textbf{Discrete}
approaches to image synthesis utilize discrete token representations as means for generative modeling. Images in pixel space are first transformed using image tokenizers ~\cite{van2017neural, esser2021taming} into discrete tokens. Multiple paradigms build upon this discrete space and iteratively generate the predicted token combination, after which the tokens are decoded back to pixel space. Autoregressive  ~\cite{lee2022autoregressive, sun2024autoregressive, yu2024randomized} models draw inspiration from language modeling, treating images as a sequence of discrete tokens and generating them by iteratively predicting the next token in the sequence. MaskGITs ~\cite{chang2022maskgit, chang2023muse, ni2024revisiting} models leverage the bi-directional nature of images by enabling parallel decoding. These models simultaneously unmask a combination of discrete tokens at each forward step, improving the efficiency and speed of the generation process. Visual autoregressive ~\cite{tian2024visual} models are inspired by the coarse-to-fine approach in image synthesis and adopt next-scale prediction to predict a combination of discrete tokens from the same scale at each step. Discrete token-based generation is gradually gaining popularity and becoming a promising approach to visual synthesis, as it is more efficient in terms of inference speed, aligns and adapts to similar paradigms in natural language processing ~\cite{radford2018improving, radford2019language, touvron2023llama}, multi-modal understanding~\cite{liu2024visual, wang2024emu3, xie2024showo} and Embodied AI~\cite{wu2024ivideogpt}.

\subsection{Discrete Image Tokenizers}

Image tokenizers play a foundational role in discrete visual generation. Extensive works have been dedicated to improving performance, training stability, compressive representation and scalability. The foundational work of VQVAE~\cite{van2017neural} and VQVAE2~\cite{razavi2019generating} introduced the paradigm of neural discrete representation learning, employing straight-through estimators to train quantized models in an end-to-end framework. Building upon this, VQGAN~\cite{esser2021taming} refines the training recipe of VQVAE, improving the fidelity of detailed image reconstruction and enabling high-resolution image generation. RQ-VAE~\cite{lee2022autoregressive} further proposes residual quantization to reduce quantization error caused by discrete tokenization. FSQ~\cite{mentzerfinite} and LFQ~\cite{yulanguage, luo2024open} investigate the impact of embedding-free tokenization and scaling codebook size to at most $2^{18}$. VQGAN-LC~\cite{zhu2024scaling} addresses the challenges in scaling codebook sizes by incorporating both pretrained vision features as codebook entries and training projectors to mitigate low utilization rates and mode collapse. IBQ~\cite{shi2024taming} introduces joint optimization of all codebook entries during each forward-backward pass, further enhancing performance with large codebook sizes. In a parallel research trajectory, SEED~\cite{ge2023planting} and TiTok~\cite{yu2024image} explore 1D tokenization, which facilitate high compression ratios.

\section{Preliminaries of Discrete Vector Quantization}
\label{sec:preliminaries}

In this section, we give a preliminary overview of discrete vector quantization. Vector quantization is a classical method which aims to project and compress visual signals into a discrete latent token space formulated by a fixed codebook $\mathcal C \in \mathbb R^{n \times d}$, where $n$ and $d$ are the size and dimension, respectively. A typical vector quantized tokenizer consists of three modules: an encoder $\mathcal E$, decoder $\mathcal D$ and quantizer $\mathcal Q$. For a given image $\mathcal I \in \mathbb R^{3 \times H \times W}$, it is first encoded by the encoder $\mathcal E$ into a compressed feature map $f \in \mathbb R^{h \times w \times d}$, then projected by $\mathcal Q$ into $z$ in the discrete latent space spanned by $\mathcal C$. Finally, $z$ is decoded back to pixel space using $\mathcal D$.

As the key component for discretization, the vector quantizer $\mathcal Q$ typically assigns features to codes by selecting the nearest neighbor $z$ based on Euclidean distance in $\mathbb R^d$ space.
\begin{equation}
    z = C_k \in \mathbb R^d, k = \argmin_{c \in C} \lVert f - c \rVert
\end{equation}

During training, the non-differentiable $\argmin$ operator poses a challenge in estimating gradients. To circumvent this, the straight-through estimator (STE)~\cite{bengio2013estimating} is employed to back-propagate the gradients back to the encoder, ensuring that the encoder and selected code are updated simultaneously.
\begin{equation}
    z_{\mathrm{quant}} = f + \mathrm{sg}[z - f]
\end{equation}

where $\mathrm{sg}[]$ is the stop-gradient operation.

This basic quantization approach has several limitations: Simultaneously updating the encoder, decoder, and quantizer leads to unstable training and makes the model highly sensitive to hyper-parameters and restart schemes, as the encoder, decoder and codebook must adapt to each other’s evolving distributions~\cite{yu2021vector, baykal2024edvae, zhustabilize}. Assigning codes based solely on Euclidean distance can cause codebook collapse and inefficient code utilization as the codebook size increases~\cite{fifty2024restructuring, zhu2024scaling, huh2023straightening}. Furthermore, the training process requires considerable time and is slow to converge~\cite{webermaskbit}.

\section{\ours Tokenizer}
\label{sec:approach}

Given a continuous VAE, a straightforward solution for converting it into a discrete tokenizer is to directly employ well-established quantization methods, \eg vector quantization (VQ) to its latent space.
Specifically, as discussed in Section~\ref{sec:preliminaries}, a set of codebook embeddings can be initialized and optimized to approximate the continuous latents at their best.
However, this naive approximation incurs significant performance drop compared to the original continuous VAE, as shown in~\cref{tab:design_analysis}.
In this section, we carefully inspect the difficulties in discretizing a continuous VAE, and present corresponding solutions step by step.

\begin{table}
\centering
    \centering
    \footnotesize
    \setlength{\tabcolsep}{2.5pt}
    \begin{tabular}{lccccc}
    \toprule
    Method & Quant. Err$\downarrow$ & rFID$\downarrow$ & PSNR$\uparrow$ & SSIM$\uparrow$ & Util.$\uparrow$ \\
    \midrule
    \textcolor{gray}{Continuous VAE~\cite{li2024autoregressive}} & \textcolor{gray}{-} & \textcolor{gray}{$0.69$} & \textcolor{gray}{$24.9$} & \textcolor{gray}{$0.716$} & \textcolor{gray}{-} \\

    \midrule

    Vector Quant. & $0.181$ & $41.52$ & $18.9$ & $0.461$ & $49\%$ \\
    + Residual Quant. & $0.094$ & $14.23$ & $20.6$ & $0.533$ & $51\%$ \\
    + Attention-based Quant.  & $0.065$ & $7.06$ & $21.4$ & $0.571$ & $100\%$ \\
    + Adapt VAE & $0.001$ & $1.34$ & $22.2$ & $0.602$ & $100\%$ \\
    \bottomrule
    \end{tabular}
    
    \vspace{-2mm}
    \caption{\textbf{Summary of our main ablation results in building the \ours tokenizer.}
    We start from an off-the-shelf continuous VAE~\cite{li2024autoregressive} (marked in \textcolor{gray}{gray}) and progressively introduce design components for effective tokenization.
    The codebook size is fixed to 65536.
    Quant: quantization. Util: utilization of codebook.
    }
    \label{tab:design_analysis}
\end{table}

\begin{figure*}[t]
    \centering
    \includegraphics[width=\linewidth]{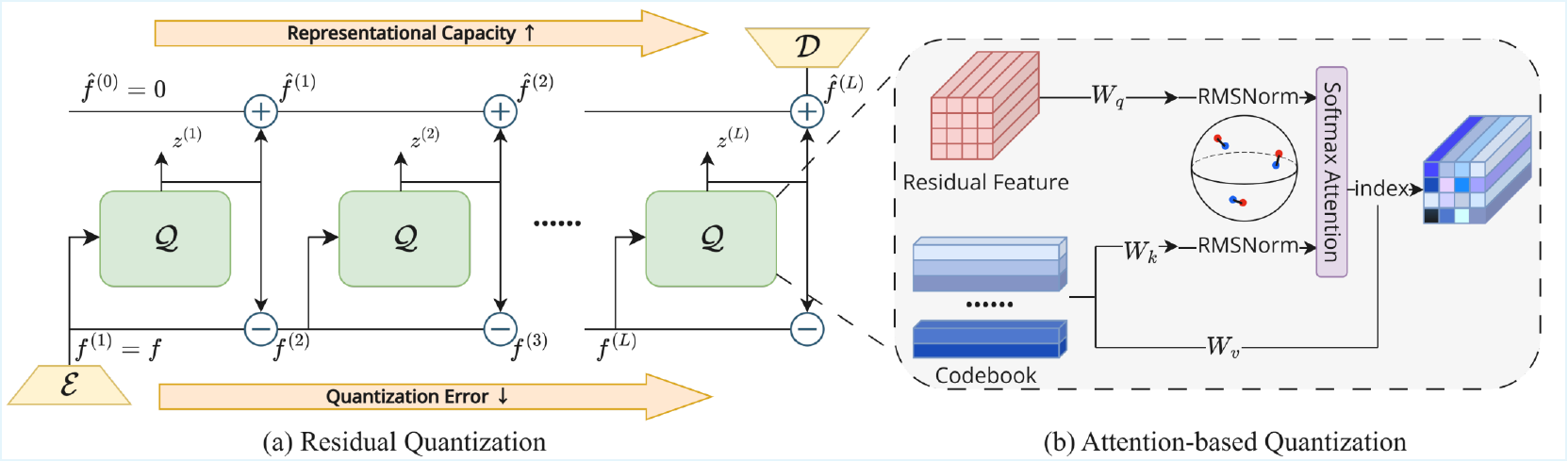}
    \caption{\textbf{Illustration of our \ours tokenizer.}
    (a) a residual quantization process of $L$ levels iteratively refines the approximation of a continuous VAE vector $f$ through a composite of multiple quantization layers, thus progressively minimizing the quantization error. Meanwhile, as the continuous VAE vector is approximated by a \emph{combination} of $L$ discrete codes, the representational capacity is significantly enlarged.
    (b) the attention-based quantization process frames discretization as a \emph{retrieval} task. Continuous features and codebook embeddings are projected and normalized onto a normed hypersphere, where the softmax attention matrix is computed to determine the confidence of code selection. As codes compete within the softmax attention framework, this approach ensures a \emph{sparse} and \emph{unambiguous} assignment.
    }
    \label{fig:tokenizer_structure}
\end{figure*}

\begin{figure}
    \centering
    \includegraphics[width=\linewidth]{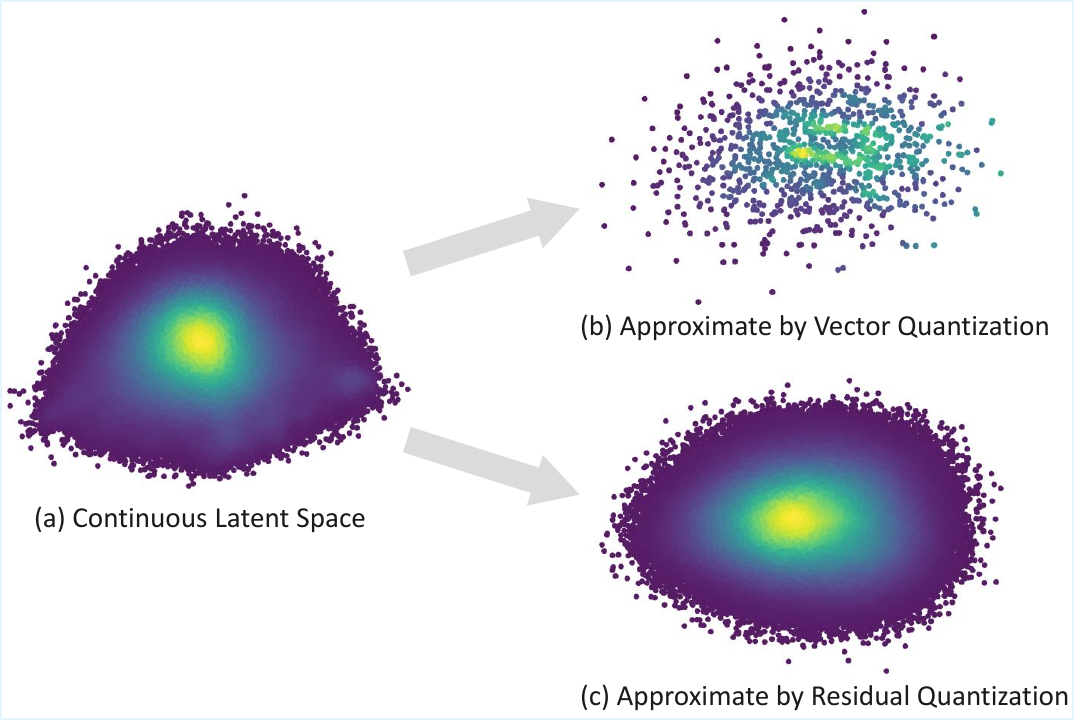}
    \caption{\textbf{Visualization of latent space approximation}: (a) the original latent space of the continuous VAE, (b) latent space approximated by vector quantization and (c) latent space approximated by residual quantization.}
    \vspace{-5mm}
    \label{fig:latent_vis}
\end{figure}

\paragraph{Problem I: insufficient representational capacity.}
To analyze the reason behind the performance drop, we first visualize the continuous latent space of the pretrained VAE and the discretized space after applying vector quantization in~\cref{fig:latent_vis} (a) and (b). Results indicate a substantial information loss during the VQ approximation process. More specifically, continuous features span across large domains in the latent space, covering not only high-density areas in the center but also marginal regions that are less populated. In contrast, discrete features quantized by VQ occupy only sparse points in the latent space. This outcome is not surprising, though, as the number of features that VQ can represent is inherently limited by the number of code embeddings in codebook. 
Though we have already employed an extensive codebook size of $65536$, exceeding that of many representative VQ tokenizers~\cite{esser2021taming, yu2021vector, sun2024autoregressive, webermaskbit}, the representational capacity still remains far from sufficient to capture the full diversity of the densely populated continuous latent space.

\paragraph{Solution: residual quantization.}
To overcome the representational limitations inherent in vector quantization, we draw inspiration from similar tasks that approximate continuous functions in numerical analysis. A common strategy to enhance the representational capacity of estimations involves decomposing the target function into the combination of multiple basis functions. For instance, $k$-th spline interpolation decomposes a continuous function $f(x)$ and represent it using $k + 1$ discrete coefficients, where increasing the level $k$ corresponds to enhanced representation capacity, allowing for more precise approximations of the original function:
\begin{equation}
    f_i(x) \approx \sum \limits_{l = 0}^k z_{i, l} (x - t_i)^l, x \in [t_i, t_{i + 1}]
\end{equation}

Inspired by this, we employ a conceptually similar technique, namely residual quantization~\cite{lee2022autoregressive} in place of vector quantization.
Residual quantization progressively refines the approximation of a continuous VAE vector $f$ through a composite of multiple quantization layers, with each layer building upon the residual of the previous one:

\begin{equation}
    \varepsilon_{l + 1} = \varepsilon_{l} - z^{(l)}
\end{equation}
where $\varepsilon_l$ and $z^{(l)}$ represents the approximation error and discrete code values at the $l$-th layer. In this way, quantization error can be iteratively minimized.
At the same time, since the continuous vector is now represented by a \emph{combination} of discrete tokens at $L$ levels:
\begin{equation}
    f \approx \sum \limits_{l = 1}^L  z^{(l)}
\end{equation}
the number of possible combinations grows exponentially with more quantization levels, thus providing a significantly larger representational capacity.

\begin{figure}
    \includegraphics[width=\linewidth]{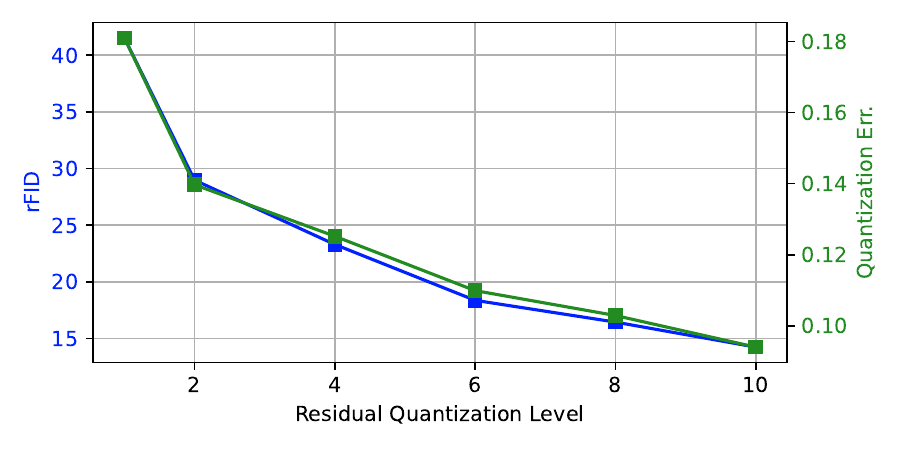}
    \vspace{-9mm}
    \caption{
    \textbf{Effect of residual quantization levels on tokenizer performance.}
    With more levels of residual quantization, quantization error is consistently minimized, and the reconstruction performance (measured by rFID) steadily improves.}
    \label{fig:err_vs_rfid}
\end{figure}

We explore the effect of residual quantization on approximating a continuous latent manifold. As shown in~\cref{fig:latent_vis}, the latent space generated by residual quantization closely resembles that of continuous latent spaces, suggesting an increased representational capacity.
As shown in~\cref{fig:err_vs_rfid}, the quantization process progressively refines at each level, leading to a consistent reduction in quantization error, which corresponds to a similar decrease in rFID.
Finally, when the number of quantization levels reaches $10$, the quantization error drops from $0.181$ to $0.094$, while rFID improves significantly from $41.52$ to $14.23$, as reported in~\cref{tab:design_analysis}.

\paragraph{Problem II: ambiguous code assignment.}
Although residual quantization has significantly improved the precision of latent space approximation, thereby enhancing reconstruction performance, we find that codebook utilization remains insufficient. As shown in~\cref{tab:design_analysis}, nearly half of the codes remain unused during quantization. To further investigate this, we randomly select $16$ continuous VAE features and for each, we visualize the Euclidean distance of its nearest $50$ neighbors from left to right. Results reveal a clear pattern of \emph{ambiguity} in code assignment. Specifically, for some features, the assignment choice has minimal impact, as these features are similarly related to multiple code vectors, leading to effective approximation regardless of assignment. Conversely, other features are far from all codes, resulting in high quantization error irrespective of assignment choice. Consequently, the imbalance in code distribution contributes to insufficient utilization of the codebook.

\begin{figure}
    \centering
    \begin{subfigure}[b]{0.48\textwidth}
        \centering
        \includegraphics[width=\linewidth]{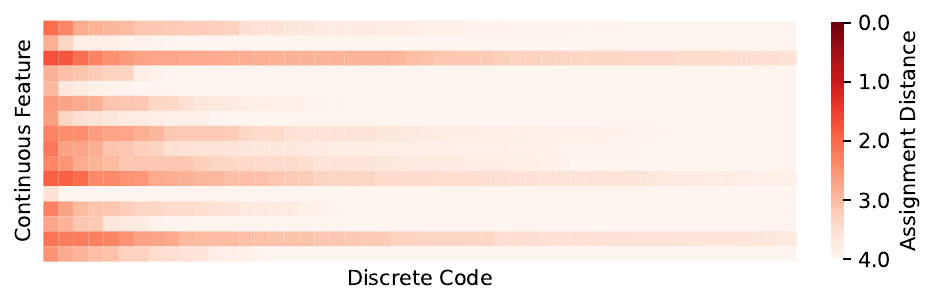}
        \caption{Assignment confidence heatmap for vector quantization}
        \label{fig:assignment_vq_heatmap}
    \end{subfigure}
    \hfill
    \begin{subfigure}[b]{0.48\textwidth}
        \centering
        \includegraphics[width=\linewidth]{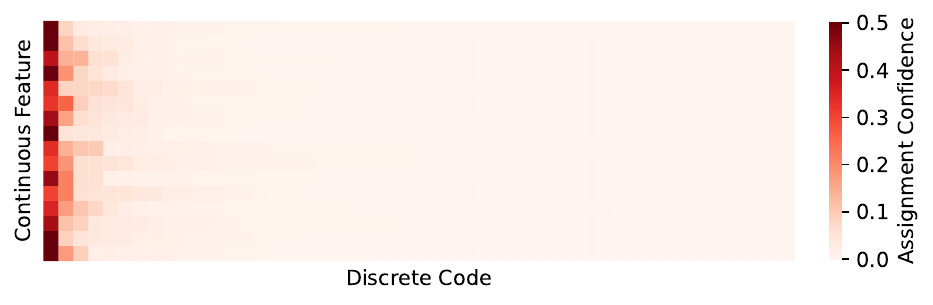}
        \caption{Assignment confidence heatmap for attention quantization}
        \label{fig:assignment_aq_heatmap}
    \end{subfigure}
    \caption{\textbf{Visualization of top assignment confidence scores}  for 16 randomly selected continuous VAE features.
    For vector quantization, we visualize the distance of codes to the continuous feature, with lower distance representing higher confidence.
    }
    \label{fig:assignment_heatmap}
\end{figure}

\begin{figure}
    \centering
    \includegraphics[width=\linewidth]{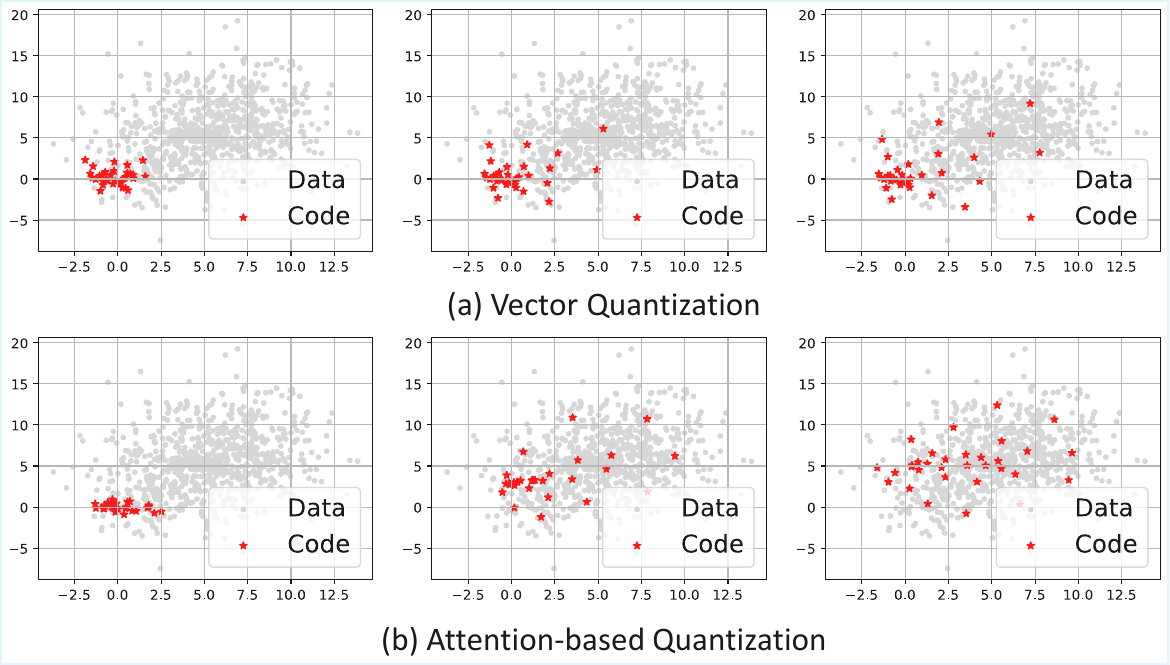}
    \caption{\textbf{Visualization of training dynamics}. In attention quantization, codes are \textit{pushed} to fully occupy the latent space, whereas vector quantization shows limited coverage of latent space.}
    \label{fig:training_dynamics}
\end{figure}

\paragraph{Solution: attention-based quantization.}
We hypothesize that this ambiguity arises from the lack of \textbf{\textit{sparsity}} in the quantization process: Ideally, \textit{sparse} assignment schemes ensure that a continuous feature is strongly associated with a single discrete code with high confidence, while remaining weakly associated to all other codes with low confidence. However, in VQ based approaches, assignments are not sparse, as features are often closely related to multiple codes in Euclidean space. This lack of sparsity makes the assignment highly sensitive to small variations in distance, resulting in ambiguous assignment. During training, the selected codebook indices are simply \textit{pulled} toward their respective clustering centroids. However, due to the lack of sparsity, multiple codes may be drawn toward the same centroid, resulting in under-utilization of the codebook. As a result, codes tend to converge around limited centers, rather than adequately covering the entire latent space. Visualization of vector quantization training dynamics, shown in~\cref{fig:training_dynamics} (a), confirms this process, where only a few codes are spread across the latent space while others form a single clique.

Motivated by these observations, we propose a novel quantization mechanism that promotes sparse assignment, and further reduces quantization errors at each quantization level. Drawing inspiration from the enhanced sparsity observed in softmax attention schemes, we design a learnable attention-based assignment strategy, as illustrated in~\cref{fig:tokenizer_structure} (b). Given an encoded feature $\mathbf F$ and codebook $\mathbf C$, the quantizer maps them to a distance space using weights $\mathbf{W}_q$ and $\mathbf{W}_k$, respectively. The projected weights are then normalized using RMSNorm~\cite{zhang2019root}, which projects the features and clusters onto a normed hypersphere~\cite{loshchilov2024ngpt}, and multiplied to compute the distance matrix $\mathbf A$ using softmax attention.
\begin{equation}
    \mathbf Q = \mathrm{rms\_norm}(\mathbf  F \mathbf W_q) \quad \mathbf K =\mathrm{rms\_norm}(\mathbf  C \mathbf W_k)
\end{equation}
\begin{equation}
    \mathbf A = \mathrm{softmax}(\frac{\mathbf Q \mathbf K^T}{\sqrt d})
\end{equation}
where $d$ is the hidden dimension. Feature at each position is assigned to the index with largest attention (\ie similarity or distance) in normed space. The assignment is propagated to the codebook weights mapped by projection $\mathbf{W}_v$:
\begin{equation}
    \hat{\mathrm F} = \mathrm{one\_hot}(\mathbf A)^T (\mathbf C \mathbf W_v)
\end{equation}
This design introduces a learnable mechanism for assigning features to discrete codes with enhanced sparsity and improved training dynamics. Attention-based quantization frames discretization as a \textit{retrieval task}, where the quantizer learns to retrieve the most appropriate index from a large codebook. Under this framework, different codes compete and suppress each other when calculating softmax attention scores, enforcing sparse assignment. This is confirmed by the visualization of confidence in~\cref{fig:assignment_aq_heatmap} (b). Additionally, this approach delivers a \textit{push} effect along with \textit{pull} during training: the chosen code index is \textit{pulled} towards the clustering centroids, reducing quantization error, while irrelevant codes are \textit{pushed} away to other regions of the latent space. This new dynamic ensures more effective codebook usage and comprehensive coverage of the latent space, as visualization of training dynamics in~\cref{fig:training_dynamics} (b) illustrate. Quantitative results in~\cref{tab:design_analysis} shows that our the attention-based assignment achieves a remarkable \emph{100\%} codebook utilization, and significantly improves the reconstruction FID from $14.23$ to $7.06$.

\begin{table*}[t]
    \centering
    \footnotesize
    \renewcommand\arraystretch{1.1}
    \resizebox{.82\linewidth}{!}{
    \begin{tabular}{lccccccc}
    \toprule
    Method & Ratio & Codebook Size & rFID $\downarrow$ & PSNR $\uparrow$ & SSIM $\uparrow$ & Utilization ($\%$) & Training (iter) \\
    \midrule
    VQGAN~\cite{esser2021taming} & $16$ & $1024$ & $7.94$ & $19.4$ & $0.500$ & $44$ & $300$K \\
    VQGAN~\cite{esser2021taming} & $16$ & $16384$ & $4.98$ & $19.9$ & $0.510$ & $5.9$ & $300$K \\
    DF-VQGAN~\cite{ni2023nuwa} & $16$ & $12288$ & $5.16$ & - & - & - & -  \\
    DQVAE~\cite{huang2023towards} & $16$ & $1024$ & $4.08$ & - & - & - & $250$K  \\
    DiVAE~\cite{shi2022divae} & $16$ & $16384$ & $4.07$ & - & - & - & -  \\
    VQGAN-LC ~\cite{zhu2024scaling} & $16$ & $16384$ & $3.01$ & $23.2$ & $0.564$ & $99$ & $100$K \\
    VQGAN-LC ~\cite{zhu2024scaling} & $16$ & $100000$ & $2.62$ & $23.8$ & $0.589$ & $99$ & $100$K \\
    LlamaGen ~\cite{sun2024autoregressive} & $16$ & $16384$  & $2.19$ & $20.8$ & $0.675$ & $97$ & $200$K \\    RQVAE~\cite{lee2022autoregressive} & - & $16384$ & $2.69$ & - & - & - & $250$K \\
    BAE~\cite{wang2023binary} & $16$ & $65536$ & $3.32$ & - & - & - & $2000$K \\
    MaskBit~\cite{webermaskbit} & $16$ & $16384$ & $1.61$ & - & - & - & $1350$K \\
    \midrule
    \ours MAR & $16$ & $16384$ & $1.43$ & $22.0$ & $0.594$ & $100$ & $\textbf{50}$\textbf{K} \\
    \ours MAR & $16$ & $65536$ & $\textbf{1.34}$ & $22.2$ & $0.602$ & $100$ & $\textbf{50}$\textbf{K} \\

    \hdashline
    \textcolor{gray}{MAR~\cite{li2024autoregressive} VAE} & \textcolor{gray}{$16$} & \textcolor{gray}{-} & \textcolor{gray}{$0.69$} & \textcolor{gray}{$24.9$} & \textcolor{gray}{$0.716$} & \textcolor{gray}{-} & \textcolor{gray}{-} \\

    \midrule[1.2pt]

    VQGAN$^{\dagger}$~\cite{esser2021taming} & $8$ & $16384$ & $1.14$ & $23.4$ & $0.670$ & $5.4$ & - \\
    ViT-VQGAN~\cite{yu2021vector} & $8$ & $8192$ & $1.28$ & - & - & - & - \\
    DF-VQGAN~\cite{ni2023nuwa} & $8$ & $8192$ & $1.38$ & - & - & - & -  \\
    DiVAE~\cite{shi2022divae} & $8$ & $16384$ & $1.28$ & - & - & - & -  \\
    OmniTokenizer$^{\dagger}$~\cite{wang2024omnitokenizer} & $8$ & $8192$ & $1.11$ & $24.0$ & $0.752$ & $100$ & - \\

    \midrule
    
    \ours FLUX & $8$ & $65536$ & $\textbf{0.43}$ & $\textbf{25.9}$ & $\textbf{0.771}$ & $100$ & $\textbf{50}$\textbf{K} \\
    
    \hdashline
    \textcolor{gray}{FLUX~\cite{flux} VAE} & \textcolor{gray}{$8$} & \textcolor{gray}{-} & \textcolor{gray}{$0.17$} & \textcolor{gray}{$31.1$} & \textcolor{gray}{$0.903$} & \textcolor{gray}{-} & \textcolor{gray}{-} \\

    \bottomrule
    \end{tabular}
    }
    \caption{\textbf{Quantitative reconstruction results on ImageNet $256 \times 256$.} Training iterations are uniformly converted and measured under global batch size of 256 following ~\cite{webermaskbit}. Ratio represents compression rate between image and latent resolution. Results from referenced continuous VAEs are marked in \textcolor{gray}{gray}. $^\dagger$ denotes training enhanced by additional web-scale data other than ImageNet.}
    \label{tab:tokenizer}
\end{table*}

\paragraph{Adapting VAE to the learned discrete space.}
Beyond approximating the continuous latent space with discrete codes, we find that slightly adjusting the original VAE parameters to accommodate the distribution shift induced by discretization is also beneficial. Specifically, we incorporate LoRA~\cite{hulora} parameters into the continuous VAE, allowing it to evolve in tandem with the learning of the discretized latent space. As shown in \cref{tab:ablation_enc_dec}, this adaptation reduces quantization error to $0.001$ and enhances the reconstruction quality, measured by rFID, to $1.34$. This significantly bridges the gap with the original continuous VAE, which achieves an rFID of $0.69$.

\paragraph{Implementation details.}
\label{subsec:training_recipe}
We follow the standard VQGAN~\cite{esser2021taming} design and train our tokenizer with a mixed combination of losses:
\begin{equation}
    \mathcal L = \mathcal L_{\mathrm{rec}} + \lambda_p \mathcal L_{p} + \lambda_{q} \mathcal L_{q} + \lambda_{\mathrm{adv}} \mathcal L_{\mathrm{adv}} + \lambda_{e} \mathcal L_{e}
\end{equation}
where $\mathcal L_{\mathrm{rec}}$ is the pixel-wise reconstruction loss, $\mathcal L_{p}$ is the LPIPS~\cite{zhang2018unreasonable} perceptual loss, $\mathcal L_{\mathrm{adv}}$ is an adversarial GAN loss passed through a StyleGAN-T discriminator~\cite{sauer2023stylegan}, $\mathcal L_{e}$ is an regularization entropy penalty for encouraging codebook usage. $\mathcal L_{q}$ is a combination of soft and hard commitment and quantization loss proposed by ~\cite{shi2024taming}:
\begin{align}
    \mathcal L_q = &\lVert \mathrm{sg}[z_{\mathrm{hard}}] - f \rVert^2 + \beta \lVert z_{\mathrm{hard}} - \mathrm{sg}[f] \rVert^2 \\
    + &\lVert z_{\mathrm{soft}} - f \rVert^2
\end{align}
We calculate the quantization loss $\mathcal L_q$ and entropy loss $\mathcal L_e$ at each level, encouraging codes at each level to minimize the quantization error and boosting codebook utilization.

\paragraph{Token-based generation.}
To verify the effectiveness of our improved tokenizer, we integrate it with the representative token-based visual generation framework MaskGIT~\cite{chang2022maskgit}. MaskGIT operates on a sequence of tokens $\mathbf{z}$ compressed by a quantized autoencoder. During training, MaskGIT randomly mask out a set of tokens using the special mask token $\mathtt{[MASK]}$. Based on the unmasked set $\mathbf{\bar{M}}$, the model predicts the logits for masked token set $\mathbf{M}$, then optimizes a BERT-style~\cite{kenton2019bert} Masked Language Modeling (MLM) Loss
\begin{equation}
    \mathcal L_{\mathrm{MLM}} = - \sum \limits_{i \in [1, N], m_i = 1} \log p(z_i | z_{\bar{\mathbf{M}}})
\end{equation}
where $p(z_i | z_{\bar{\mathbf{M}}})$ is the predicted logits of token index $z_i$ based on the unmasked token set $\bar M$.

During inference, MaskGIT start from a fully masked sequence and decodes iteratively. During each step, the model predicts logits for all masked positions and samples a portion of most confident tokens to be unmasked according to a predefined schedule. The final fully unmasked sequence is then decoded back to pixel space using the tokenizer decoder.

\section{Experiments}
\label{sec:exp}

\subsection{Experiment Setup}
\label{subsec:experiment_setup}

\paragraph{Datasets and evaluation metrics.} We conduct our experiments on the ImageNet dataset~\cite{deng2009imagenet}, specifically utilizing images at a resolution of $256 \times 256$. Our study involves both training our tokenizer and evaluating generative models using this dataset. To assess the reconstruction quality of our approach, we employ multiple evaluation metrics, including reconstruction FID (rFID), Peak Signal-to-Noise Ratio (PSNR), and Structural Similarity Index Measure (SSIM). These metrics provide a comprehensive evaluation of both perceptual and pixel-level reconstruction accuracy.   

\vspace{1mm}
\paragraph{Tokenizer training.} Leveraging the continuous latent space of pretrained VAEs, we utilize and freeze the encoder-decoder weights from two standard continuous VAE: a $16 \times$ MAR~\cite{li2024autoregressive} VAE and a $8 \times$ FLUX VAE~\cite{flux}. To adapt the continuous encoder-decoder to discrete codes, we implement LoRA~\cite{hulora} modules exclusively on the convolution blocks and finetune the added modules along with the quantizer during training. In practice, we adopt a LoRA rank of $32$ for both encoder and decoder. We follow a main stream implementation of residual quantization introduced in ~\cite{tian2024visual} and set the quantization level to $10$. All models are trained with an accumulated batch size of $512$ for $10$ epochs on the ImageNet $256 \times 256$ dataset with a $5e-4$ learning rate.

\vspace{1mm}
\paragraph{Generative model training.} MaskGIT models are trained for $500$K steps with an accumulated batch size of $2048$ on the ImageNet $256 \times 256$ dataset with $4e-4$ learning rate. In addition to training on the standard tokenized representation, we introduce learnable positional embeddings specific to each quantization level. This modification enhances the model’s ability to differentiate between tokens from different levels of the residual quantization hierarchy, ultimately improving the decoding process. 

\subsection{Main Results}

\begin{figure*}[t]
\centering
{\includegraphics[width=\textwidth]{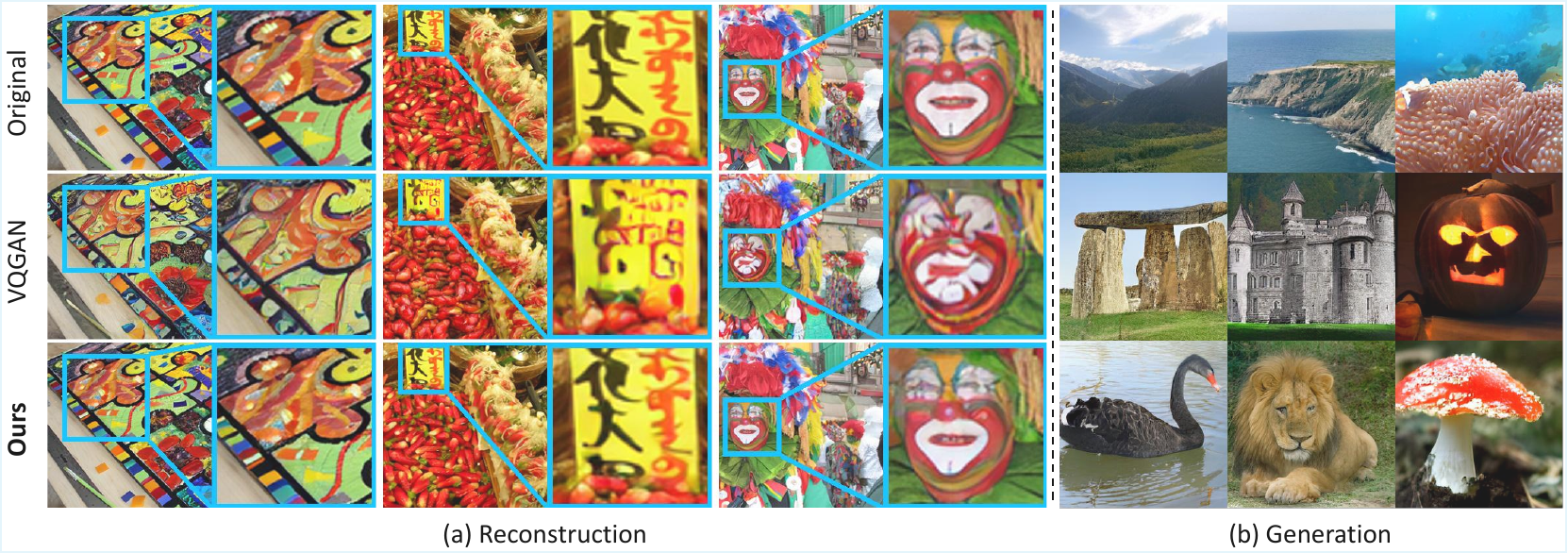}}
\caption{\textbf{Visualization of samples on ImageNet $256 \times 256$}. (a) Reconstruction results by \ours tokenizer. Compared with VQGAN, \ours showcases higher fidelity and effectively preserves rich details. (b) Generated samples by combining \ours with MaskGIT.}
\label{fig:visualization}
\end{figure*}

\paragraph{Tokenizer performance.} 
We present the reconstruction results in~\cref{tab:tokenizer}. There are three key observations from our findings: First, our \ours tokenizer achieves competitive performance with $0.43$ and $1.34$ rFID in $8 \times$ and $16 \times$ reduction settings, respectively, significantly outperforming corresponding baselines in terms of reconstruction quality and fidelity. Second, our tokenizer achieves a high utilization rate of $100\%$ with a large codebook of $65536$, compared to VQGAN’s utilization rate of $5.9\%$ with $16384$ codes. This demonstrates that the combined approach of training in a pretrained continuous latent space and employing attention-based sparse quantization mechanisms prevents codebook collapse and encourages high utilization, providing the potential for scaling to even larger scopes. Finally, our approach requires substantially less training computation than all baseline methods, achieving a $6 \times$ speedup compared to the representative VQGAN training recipe. Additional qualitative visualizations and comparisons are provided in~\cref{fig:visualization} (a).

\begin{table}[h]
\centering

    \centering
    \setlength{\tabcolsep}{6pt}
    \footnotesize
\resizebox{\linewidth}{!}{
    \begin{tabular}{lcccc}
    \toprule
    Method & Type & \#Params & Steps  & FID$\downarrow$  \\
    \midrule
    ADM~\cite{dhariwal2021diffusion} & Diff. & $554$M & $250$ & $10.94$ \\
    LDM-4-G~\cite{rombach2022high} & Diff. & $400$M & $250$ & $3.60$ \\
    DiT-L/2~\cite{peebles2023scalable} & Diff. & $458$M & $250$ & $5.02$ \\
    UViT-L/2~\cite{bao2023all} & Diff. & $287$M & $50$ & $3.40$ \\

    LDM-4-G$^{\dagger}$~\cite{rombach2022high} & Diff. & $400$M & $8$ & $4.56$ \\
    DiT-XL/2$^{\dagger}$~\cite{peebles2023scalable} & Diff. & $675$M & $8$ & $5.18$ \\
    
    RQVAE-GPT~\cite{lee2022autoregressive} & AR & $480$M & $256$ & $15.7$ \\
    VQGAN-LC-GPT~\cite{zhu2024scaling} & AR & $404$M & $256$ & $15.4$ \\
    ViT-VQGAN-GPT~\cite{yu2021vector} & AR & $650$M & $1024$ & $8.81$ \\
    RQ-Transformer~\cite{lee2022autoregressive} & AR & $3.8$B & $64$ & $7.55$ \\

    LlamaGen-L-384~\cite{sun2024autoregressive} & AR & $343$M & $576$ & $3.07$ \\

    VAR~\cite{tian2024visual} & VAR & $310$M & $10$ & $3.30$ \\

    MaskGIT~\cite{chang2022maskgit} & Mask. & $227$M & $12$ & $4.92$ \\
    MaskGIT-FSQ~\cite{mentzerfinite} & Mask. & $225$M & $12$ & $4.53$ \\
    MAGE~\cite{li2023mage} & Mask. & $230$M & $20$ & $6.93$ \\
    ENAT~\cite{nienat} & Mask. & $219$M & $8$ & $3.53$ \\
    \midrule
    \ours MaskGIT-L & Mask. & $195$M & $8$ & $3.17$ \\
    \ours MaskGIT-L & Mask. & $195$M & $32$ & $\textbf{2.66}$ \\
    \bottomrule
    \end{tabular}
}
    \caption{\textbf{Quantitative generation results on ImageNet $256 \times 256$.} "-384" denotes images generated at $384$ resolution and resized back to $256$ during evaluation. $^{\dagger}$ denotes diffusion schedule augmented by DPM-Solver~\cite{lu2022dpm}.}
    \label{tab:generation}
\end{table}

\begin{table*}
    \centering
    \begin{subtable}{0.32\textwidth}
        \vtop{
            \centering
            \setlength{\tabcolsep}{4pt}
            \footnotesize
            \begin{tabular}{ccccc}
            \toprule
            Enc. & Dec. & rFID$\downarrow$ & PSNR$\uparrow$ & SSIM$\uparrow$ \\
            \midrule
            \xmark & \xmark & $7.06$ & $21.4$ & $0.571$ \\
            \cmark & \xmark & $5.75$ & $21.3$ & $0.551$ \\
            \xmark & \cmark & $1.59$ & $22.1$ & $0.604$ \\
            \default{\cmark} & \default{\cmark} & \default{$1.34$} & \default{$22.2$} & \default{$0.602$} \\
            \bottomrule
            \end{tabular}
        }
        \caption{Comparison on effect of tuned weights.}
        \label{tab:ablation_enc_dec}
    \end{subtable}
    \begin{subtable}{0.32\textwidth}
        \vtop{
            \centering
            \setlength{\tabcolsep}{4pt}
            \footnotesize
            \begin{tabular}{ccccc}
            \toprule
            Codebook Size & rFID$\downarrow$ & PSNR$\uparrow$ & SSIM$\uparrow$  \\
            \midrule
            $1024$ & $2.07$ & $21.2$ & $0.561$ \\
            $4096$ & $1.76$ & $21.7$ & $0.582$ \\
            $16384$ & $1.43$ & $22.0$ & $0.594$ \\
            \default{$65536$} & \default{$1.34$} & \default{$22.2$} & \default{$0.602$} \\
            \bottomrule
            \end{tabular}
        }
        \caption{Comparison on effect of codebook size.}
        \label{tab:effect_codebook_size}
    \end{subtable}
    \begin{subtable}{0.32\textwidth}
        \vtop{
            \centering
            \setlength{\tabcolsep}{4pt}
            \footnotesize
            \begin{tabular}{ccccc}
            \toprule
            Norm & rFID$\downarrow$ & PSNR$\uparrow$ & SSIM$\uparrow$  \\
            \midrule
            w/o & $1.87$ & $21.51$ & $0.570$ \\
            LayerNorm & $1.37$ & $22.1$ & $0.600$  \\
            \default{RMSNorm} & \default{$1.34$} & \default{$22.2$} & \default{$0.602$} \\
            \bottomrule
            \end{tabular}
            }
        \vspace{9.5pt}
        \caption{Comparison on effect of normalization.}
        \label{tab:effect_norm}
    \end{subtable}
    \caption{\textbf{Ablation studies.}
    We use CODA-MAR tokenizer and mark our default setting in \colorbox{defaultcolor}{gray}}
    \vspace{-5mm}
\end{table*}

\vspace{1mm}
\paragraph{Image generation.}
In~\cref{tab:generation}, We compare our approach with leading image generation paradigms, including continuous diffusion-based methods, discrete MaskGIT, autoregressive models, and visual autoregressive models. Our key observations are as follows: 1) Compared to other MaskGIT-based models, our approach achieves a lower FID of $3.17$, compared to $4.92$ for MaskGIT and $4.53$ for MaskGIT-FSQ, demonstrating enhanced generation quality delivered by a more effective tokenization strategy. 2) Our model demonstrates strong overall performance relative to other main stream generation paradigms, with an FID of $2.66$ compared to $3.60$ for LDM and $3.07$ for LlamaGen. 3) Our approach achieves high-quality generation results comparable to continuous methods while maintaining the superior efficiency of discrete models, requiring fewer sampling steps. Additional quantitative text-to-image generation results and qualitative results of generated images from our approach are demonstrated in ~\cref{tab:mscoco} and ~\cref{fig:visualization} (b) respectively.

\begin{table}[h]
    \centering
    \setlength{\tabcolsep}{2pt}
    \footnotesize
    \vspace{-1.5mm}
    \tablestyle{2mm}{1}
    \begin{tabular}{c|cccc}
    Method & VQ-Diffusion & U-Net & U-ViT-S  & \default{\ours MaskGIT-S}  \\\shline
    FID$\downarrow$ & $13.86$ & $7.32$ & $5.95$ & \default{$\textbf{5.44}$} \\
    \end{tabular}
    \caption{Text-to-image generation on MS-COCO.}
    \label{tab:mscoco}
    \vspace{-5mm}
\end{table}

\subsection{Ablation Studies}

In this section, we present more ablation studies to justify the effectiveness of our approach.

\vspace{1mm}
\paragraph{VAE parameters.} We investigate the impact of fine-tuning the encoder and decoder LoRA weights when adapting a continuous VAE into a discrete tokenizer. To this end, we implement tokenizer training with different configurations: freezing both the encoder and decoder, training LoRA modules exclusively on the encoder, training LoRA modules exclusively on the decoder, and training LoRA modules on both the encoder and decoder. The results are presented in Table~\ref{tab:ablation_enc_dec}. Interestingly, we observe that adapting the encoder and decoder affect performance differently. Specifically, 
training a small number of parameters on the decoder leads to noteworthy improvements, whereas training on the encoder yields only marginal gains. This observed asymmetry in training modules align with our motivation that approximating the discrete model closely to the continuous latent space is sufficient to achieve competitive performance.

\vspace{1mm}
\paragraph{Effect of scaling codebook size.} Results in~\cref{tab:effect_codebook_size} demonstrate that our approach enables stable scaling of the codebook size from $1024$ to $65536$, leading to a consistent improvement in tokenizer performance, with rFID steadily decreasing from $2.07$ to $1.34$. This highlights the effectiveness of our method in fully utilizing an expanded codebook to refine discrete representations without encountering common issues such as codebook collapse or under-utilization.

\vspace{1mm}
\paragraph{Effect of normalization.}
We argue that normalization used in the attention quantization modules play a positive role in better assignment, leading to improved reconstruction results. As results in~\cref{tab:effect_norm} demonstrate, projecting the query-key pairs into a normed space improves rFID from $1.87$ to $1.34$. Other types of normalization, \eg LayerNorm is also capable of achieving similar results compared to using RMSNorm.

\section{Conclusion}
\label{sec:conclusion}

In this work, we present \ours tokenizers, a novel approach for training discrete tokenizers by leveraging the pretrained space of continuous VAEs for compression, while optimizing discretization through principled adaptations. Through a series of carefully designed discretization mechanisms, \ours is able to approximate continuous latent space using discrete tokens. Experiments on ImageNet demonstrate that \ours enables improved high fidelity reconstruction performance with effective codebook utilization, while requiring only minimal training. Additional experiments reveal that \ours unlocks new potential for enhanced discrete generation.

\subsection*{Acknowledgement}

This work is supported in part by the National Key R\&D Program of China (2024YFB4708200), the National Natural Science Foundation of China (U24B20173, 62321005), and the Scientific Research Innovation Capability Support Project for Young Faculty (ZYGXQNJSKYCXNLZCXM-I20).
{
    \small
    \bibliographystyle{ieeenat_fullname}
    \bibliography{main}

\begin{thebibliography}{61}
\providecommand{\natexlab}[1]{#1}
\providecommand{\url}[1]{\texttt{#1}}
\expandafter\ifx\csname urlstyle\endcsname\relax
  \providecommand{\doi}[1]{doi: #1}\else
  \providecommand{\doi}{doi: \begingroup \urlstyle{rm}\Url}\fi

\bibitem[Achiam et~al.(2023)Achiam, Adler, Agarwal, Ahmad, Akkaya, Aleman, Almeida, Altenschmidt, Altman, Anadkat, et~al.]{achiam2023gpt}
Josh Achiam, Steven Adler, Sandhini Agarwal, Lama Ahmad, Ilge Akkaya, Florencia~Leoni Aleman, Diogo Almeida, Janko Altenschmidt, Sam Altman, Shyamal Anadkat, et~al.
\newblock Gpt-4 technical report.
\newblock \emph{arXiv preprint arXiv:2303.08774}, 2023.

\bibitem[Bao et~al.(2023)Bao, Nie, Xue, Cao, Li, Su, and Zhu]{bao2023all}
Fan Bao, Shen Nie, Kaiwen Xue, Yue Cao, Chongxuan Li, Hang Su, and Jun Zhu.
\newblock All are worth words: A vit backbone for diffusion models.
\newblock In \emph{Proceedings of the IEEE/CVF conference on computer vision and pattern recognition}, pages 22669--22679, 2023.

\bibitem[Baykal et~al.(2024)Baykal, Kandemir, and Unal]{baykal2024edvae}
Gulcin Baykal, Melih Kandemir, and Gozde Unal.
\newblock Edvae: Mitigating codebook collapse with evidential discrete variational autoencoders.
\newblock \emph{Pattern Recognition}, 156:\penalty0 110792, 2024.

\bibitem[Bengio et~al.(2013)Bengio, L{\'e}onard, and Courville]{bengio2013estimating}
Yoshua Bengio, Nicholas L{\'e}onard, and Aaron Courville.
\newblock Estimating or propagating gradients through stochastic neurons for conditional computation.
\newblock \emph{arXiv preprint arXiv:1308.3432}, 2013.

\bibitem[BlackForest(2024)]{flux}
BlackForest.
\newblock Black forest labs; frontier ai lab, 2024.

\bibitem[Brown et~al.(2020)Brown, Mann, Ryder, Subbiah, Kaplan, Dhariwal, Neelakantan, Shyam, Sastry, Askell, et~al.]{brown2020language}
Tom Brown, Benjamin Mann, Nick Ryder, Melanie Subbiah, Jared~D Kaplan, Prafulla Dhariwal, Arvind Neelakantan, Pranav Shyam, Girish Sastry, Amanda Askell, et~al.
\newblock Language models are few-shot learners.
\newblock \emph{Advances in neural information processing systems}, 33:\penalty0 1877--1901, 2020.

\bibitem[Chang et~al.(2022)Chang, Zhang, Jiang, Liu, and Freeman]{chang2022maskgit}
Huiwen Chang, Han Zhang, Lu Jiang, Ce Liu, and William~T Freeman.
\newblock Maskgit: Masked generative image transformer.
\newblock In \emph{Proceedings of the IEEE/CVF Conference on Computer Vision and Pattern Recognition}, pages 11315--11325, 2022.

\bibitem[Chang et~al.(2023)Chang, Zhang, Barber, Maschinot, Lezama, Jiang, Yang, Murphy, Freeman, Rubinstein, et~al.]{chang2023muse}
Huiwen Chang, Han Zhang, Jarred Barber, AJ Maschinot, Jos{\'e} Lezama, Lu Jiang, Ming-Hsuan Yang, Kevin Murphy, William~T Freeman, Michael Rubinstein, et~al.
\newblock Muse: Text-to-image generation via masked generative transformers.
\newblock In \emph{Proceedings of the 40th International Conference on Machine Learning}, pages 4055--4075, 2023.

\bibitem[Deng et~al.(2009)Deng, Dong, Socher, Li, Li, and Fei-Fei]{deng2009imagenet}
Jia Deng, Wei Dong, Richard Socher, Li-Jia Li, Kai Li, and Li Fei-Fei.
\newblock Imagenet: A large-scale hierarchical image database.
\newblock In \emph{2009 IEEE conference on computer vision and pattern recognition}, pages 248--255. Ieee, 2009.

\bibitem[Dhariwal and Nichol(2021)]{dhariwal2021diffusion}
Prafulla Dhariwal and Alexander Nichol.
\newblock Diffusion models beat gans on image synthesis.
\newblock \emph{Advances in neural information processing systems}, 34:\penalty0 8780--8794, 2021.

\bibitem[Esser et~al.(2021)Esser, Rombach, and Ommer]{esser2021taming}
Patrick Esser, Robin Rombach, and Bjorn Ommer.
\newblock Taming transformers for high-resolution image synthesis.
\newblock In \emph{Proceedings of the IEEE/CVF conference on computer vision and pattern recognition}, pages 12873--12883, 2021.

\bibitem[Fifty et~al.(2024)Fifty, Junkins, Duan, Iger, Liu, Amid, Thrun, and R{\'e}]{fifty2024restructuring}
Christopher Fifty, Ronald~G Junkins, Dennis Duan, Aniketh Iger, Jerry~W Liu, Ehsan Amid, Sebastian Thrun, and Christopher R{\'e}.
\newblock Restructuring vector quantization with the rotation trick.
\newblock \emph{arXiv preprint arXiv:2410.06424}, 2024.

\bibitem[Ge et~al.(2023)Ge, Ge, Zeng, Wang, and Shan]{ge2023planting}
Yuying Ge, Yixiao Ge, Ziyun Zeng, Xintao Wang, and Ying Shan.
\newblock Planting a seed of vision in large language model.
\newblock \emph{arXiv preprint arXiv:2307.08041}, 2023.

\bibitem[Hu et~al.()Hu, Wallis, Allen-Zhu, Li, Wang, Wang, Chen, et~al.]{hulora}
Edward~J Hu, Phillip Wallis, Zeyuan Allen-Zhu, Yuanzhi Li, Shean Wang, Lu Wang, Weizhu Chen, et~al.
\newblock Lora: Low-rank adaptation of large language models.
\newblock In \emph{International Conference on Learning Representations}.

\bibitem[Huang et~al.(2023)Huang, Mao, Chen, and Zhang]{huang2023towards}
Mengqi Huang, Zhendong Mao, Zhuowei Chen, and Yongdong Zhang.
\newblock Towards accurate image coding: Improved autoregressive image generation with dynamic vector quantization.
\newblock In \emph{Proceedings of the IEEE/CVF Conference on Computer Vision and Pattern Recognition}, pages 22596--22605, 2023.

\bibitem[Huh et~al.(2023)Huh, Cheung, Agrawal, and Isola]{huh2023straightening}
Minyoung Huh, Brian Cheung, Pulkit Agrawal, and Phillip Isola.
\newblock Straightening out the straight-through estimator: Overcoming optimization challenges in vector quantized networks.
\newblock In \emph{International Conference on Machine Learning}, pages 14096--14113. PMLR, 2023.

\bibitem[Kenton and Toutanova(2019)]{kenton2019bert}
Jacob Devlin Ming-Wei~Chang Kenton and Lee~Kristina Toutanova.
\newblock Bert: Pre-training of deep bidirectional transformers for language understanding.
\newblock In \emph{Proceedings of naacL-HLT}. Minneapolis, Minnesota, 2019.

\bibitem[Kingma(2013)]{kingma2013auto}
Diederik~P Kingma.
\newblock Auto-encoding variational bayes.
\newblock \emph{arXiv preprint arXiv:1312.6114}, 2013.

\bibitem[Lee et~al.(2022)Lee, Kim, Kim, Cho, and Han]{lee2022autoregressive}
Doyup Lee, Chiheon Kim, Saehoon Kim, Minsu Cho, and Wook-Shin Han.
\newblock Autoregressive image generation using residual quantization.
\newblock In \emph{Proceedings of the IEEE/CVF Conference on Computer Vision and Pattern Recognition}, pages 11523--11532, 2022.

\bibitem[Li et~al.(2023)Li, Chang, Mishra, Zhang, Katabi, and Krishnan]{li2023mage}
Tianhong Li, Huiwen Chang, Shlok Mishra, Han Zhang, Dina Katabi, and Dilip Krishnan.
\newblock Mage: Masked generative encoder to unify representation learning and image synthesis.
\newblock In \emph{Proceedings of the IEEE/CVF Conference on Computer Vision and Pattern Recognition}, pages 2142--2152, 2023.

\bibitem[Li et~al.(2024)Li, Tian, Li, Deng, and He]{li2024autoregressive}
Tianhong Li, Yonglong Tian, He Li, Mingyang Deng, and Kaiming He.
\newblock Autoregressive image generation without vector quantization.
\newblock \emph{arXiv preprint arXiv:2406.11838}, 2024.

\bibitem[Liu et~al.(2024)Liu, Li, Wu, and Lee]{liu2024visual}
Haotian Liu, Chunyuan Li, Qingyang Wu, and Yong~Jae Lee.
\newblock Visual instruction tuning.
\newblock \emph{Advances in neural information processing systems}, 36, 2024.

\bibitem[Loshchilov et~al.(2024)Loshchilov, Hsieh, Sun, and Ginsburg]{loshchilov2024ngpt}
Ilya Loshchilov, Cheng-Ping Hsieh, Simeng Sun, and Boris Ginsburg.
\newblock ngpt: Normalized transformer with representation learning on the hypersphere.
\newblock \emph{arXiv preprint arXiv:2410.01131}, 2024.

\bibitem[Lu et~al.(2022)Lu, Zhou, Bao, Chen, Li, and Zhu]{lu2022dpm}
Cheng Lu, Yuhao Zhou, Fan Bao, Jianfei Chen, Chongxuan Li, and Jun Zhu.
\newblock Dpm-solver: A fast ode solver for diffusion probabilistic model sampling in around 10 steps.
\newblock \emph{Advances in Neural Information Processing Systems}, 35:\penalty0 5775--5787, 2022.

\bibitem[Luo et~al.(2024)Luo, Shi, Ge, Yang, Wang, and Shan]{luo2024open}
Zhuoyan Luo, Fengyuan Shi, Yixiao Ge, Yujiu Yang, Limin Wang, and Ying Shan.
\newblock Open-magvit2: An open-source project toward democratizing auto-regressive visual generation.
\newblock \emph{arXiv preprint arXiv:2409.04410}, 2024.

\bibitem[Mentzer et~al.()Mentzer, Minnen, Agustsson, and Tschannen]{mentzerfinite}
Fabian Mentzer, David Minnen, Eirikur Agustsson, and Michael Tschannen.
\newblock Finite scalar quantization: Vq-vae made simple.
\newblock In \emph{The Twelfth International Conference on Learning Representations}.

\bibitem[Ni et~al.(2023)Ni, Li, and Zuo]{ni2023nuwa}
Minheng Ni, Xiaoming Li, and Wangmeng Zuo.
\newblock Nuwa-lip: language-guided image inpainting with defect-free vqgan.
\newblock In \emph{Proceedings of the IEEE/CVF Conference on Computer Vision and Pattern Recognition}, pages 14183--14192, 2023.

\bibitem[Ni et~al.()Ni, Wang, Zhou, Han, Guo, Liu, Yao, and Huang]{nienat}
Zanlin Ni, Yulin Wang, Renping Zhou, Yizeng Han, Jiayi Guo, Zhiyuan Liu, Yuan Yao, and Gao Huang.
\newblock Enat: Rethinking spatial-temporal interactions in token-based image synthesis.
\newblock In \emph{The Thirty-eighth Annual Conference on Neural Information Processing Systems}.

\bibitem[Ni et~al.(2024{\natexlab{a}})Ni, Wang, Zhou, Guo, Hu, Liu, Song, Yao, and Huang]{ni2024revisiting}
Zanlin Ni, Yulin Wang, Renping Zhou, Jiayi Guo, Jinyi Hu, Zhiyuan Liu, Shiji Song, Yuan Yao, and Gao Huang.
\newblock Revisiting non-autoregressive transformers for efficient image synthesis.
\newblock In \emph{Proceedings of the IEEE/CVF Conference on Computer Vision and Pattern Recognition}, pages 7007--7016, 2024{\natexlab{a}}.

\bibitem[Ni et~al.(2024{\natexlab{b}})Ni, Wang, Zhou, Lu, Guo, Hu, Liu, Yao, and Huang]{Ni2024AdaNAT}
Zanlin Ni, Yulin Wang, Renping Zhou, Rui Lu, Jiayi Guo, Jinyi Hu, Zhiyuan Liu, Yuan Yao, and Gao Huang.
\newblock Adanat: Exploring adaptive policy for token-based image generation.
\newblock In \emph{ECCV}, 2024{\natexlab{b}}.

\bibitem[Peebles and Xie(2023)]{peebles2023scalable}
William Peebles and Saining Xie.
\newblock Scalable diffusion models with transformers.
\newblock In \emph{Proceedings of the IEEE/CVF International Conference on Computer Vision}, pages 4195--4205, 2023.

\bibitem[Podell et~al.()Podell, English, Lacey, Blattmann, Dockhorn, M{\"u}ller, Penna, and Rombach]{podellsdxl}
Dustin Podell, Zion English, Kyle Lacey, Andreas Blattmann, Tim Dockhorn, Jonas M{\"u}ller, Joe Penna, and Robin Rombach.
\newblock Sdxl: Improving latent diffusion models for high-resolution image synthesis.
\newblock In \emph{The Twelfth International Conference on Learning Representations}.

\bibitem[Radford(2018)]{radford2018improving}
Alec Radford.
\newblock Improving language understanding by generative pre-training.
\newblock 2018.

\bibitem[Radford et~al.()Radford, Wu, Child, Luan, Amodei, Sutskever, et~al.]{radford2019language}
Alec Radford, Jeffrey Wu, Rewon Child, David Luan, Dario Amodei, Ilya Sutskever, et~al.
\newblock Language models are unsupervised multitask learners.

\bibitem[Raffel et~al.(2020)Raffel, Shazeer, Roberts, Lee, Narang, Matena, Zhou, Li, and Liu]{raffel2020exploring}
Colin Raffel, Noam Shazeer, Adam Roberts, Katherine Lee, Sharan Narang, Michael Matena, Yanqi Zhou, Wei Li, and Peter~J Liu.
\newblock Exploring the limits of transfer learning with a unified text-to-text transformer.
\newblock \emph{Journal of machine learning research}, 21\penalty0 (140):\penalty0 1--67, 2020.

\bibitem[Razavi et~al.(2019)Razavi, Van~den Oord, and Vinyals]{razavi2019generating}
Ali Razavi, Aaron Van~den Oord, and Oriol Vinyals.
\newblock Generating diverse high-fidelity images with vq-vae-2.
\newblock \emph{Advances in neural information processing systems}, 32, 2019.

\bibitem[Rombach et~al.(2022)Rombach, Blattmann, Lorenz, Esser, and Ommer]{rombach2022high}
Robin Rombach, Andreas Blattmann, Dominik Lorenz, Patrick Esser, and Bj{\"o}rn Ommer.
\newblock High-resolution image synthesis with latent diffusion models.
\newblock In \emph{Proceedings of the IEEE/CVF conference on computer vision and pattern recognition}, pages 10684--10695, 2022.

\bibitem[Sauer et~al.(2023)Sauer, Karras, Laine, Geiger, and Aila]{sauer2023stylegan}
Axel Sauer, Tero Karras, Samuli Laine, Andreas Geiger, and Timo Aila.
\newblock Stylegan-t: Unlocking the power of gans for fast large-scale text-to-image synthesis.
\newblock In \emph{International conference on machine learning}, pages 30105--30118. PMLR, 2023.

\bibitem[Shi et~al.(2024)Shi, Luo, Ge, Yang, Shan, and Wang]{shi2024taming}
Fengyuan Shi, Zhuoyan Luo, Yixiao Ge, Yujiu Yang, Ying Shan, and Limin Wang.
\newblock Taming scalable visual tokenizer for autoregressive image generation.
\newblock \emph{arXiv preprint arXiv:2412.02692}, 2024.

\bibitem[Shi et~al.(2022)Shi, Wu, Liang, Liu, and Duan]{shi2022divae}
Jie Shi, Chenfei Wu, Jian Liang, Xiang Liu, and Nan Duan.
\newblock Divae: Photorealistic images synthesis with denoising diffusion decoder.
\newblock \emph{arXiv preprint arXiv:2206.00386}, 2022.

\bibitem[Sun et~al.(2024)Sun, Jiang, Chen, Zhang, Peng, Luo, and Yuan]{sun2024autoregressive}
Peize Sun, Yi Jiang, Shoufa Chen, Shilong Zhang, Bingyue Peng, Ping Luo, and Zehuan Yuan.
\newblock Autoregressive model beats diffusion: Llama for scalable image generation.
\newblock \emph{arXiv preprint arXiv:2406.06525}, 2024.

\bibitem[Tian et~al.(2024)Tian, Jiang, Yuan, Peng, and Wang]{tian2024visual}
Keyu Tian, Yi Jiang, Zehuan Yuan, Bingyue Peng, and Liwei Wang.
\newblock Visual autoregressive modeling: Scalable image generation via next-scale prediction.
\newblock \emph{arXiv preprint arXiv:2404.02905}, 2024.

\bibitem[Touvron et~al.(2023)Touvron, Lavril, Izacard, Martinet, Lachaux, Lacroix, Rozi{\`e}re, Goyal, Hambro, Azhar, et~al.]{touvron2023llama}
Hugo Touvron, Thibaut Lavril, Gautier Izacard, Xavier Martinet, Marie-Anne Lachaux, Timoth{\'e}e Lacroix, Baptiste Rozi{\`e}re, Naman Goyal, Eric Hambro, Faisal Azhar, et~al.
\newblock Llama: Open and efficient foundation language models.
\newblock \emph{arXiv preprint arXiv:2302.13971}, 2023.

\bibitem[Van Den~Oord et~al.(2017)Van Den~Oord, Vinyals, et~al.]{van2017neural}
Aaron Van Den~Oord, Oriol Vinyals, et~al.
\newblock Neural discrete representation learning.
\newblock \emph{Advances in neural information processing systems}, 30, 2017.

\bibitem[Vaswani(2017)]{vaswani2017attention}
A Vaswani.
\newblock Attention is all you need.
\newblock \emph{Advances in Neural Information Processing Systems}, 2017.

\bibitem[Wang et~al.(2024{\natexlab{a}})Wang, Jiang, Yuan, Peng, Wu, and Jiang]{wang2024omnitokenizer}
Junke Wang, Yi Jiang, Zehuan Yuan, Binyue Peng, Zuxuan Wu, and Yu-Gang Jiang.
\newblock Omnitokenizer: A joint image-video tokenizer for visual generation.
\newblock \emph{arXiv preprint arXiv:2406.09399}, 2024{\natexlab{a}}.

\bibitem[Wang et~al.(2024{\natexlab{b}})Wang, Zhang, Luo, Sun, Cui, Wang, Zhang, Wang, Li, Yu, et~al.]{wang2024emu3}
Xinlong Wang, Xiaosong Zhang, Zhengxiong Luo, Quan Sun, Yufeng Cui, Jinsheng Wang, Fan Zhang, Yueze Wang, Zhen Li, Qiying Yu, et~al.
\newblock Emu3: Next-token prediction is all you need.
\newblock \emph{arXiv preprint arXiv:2409.18869}, 2024{\natexlab{b}}.

\bibitem[Wang et~al.(2023)Wang, Wang, Liu, and Qiu]{wang2023binary}
Ze Wang, Jiang Wang, Zicheng Liu, and Qiang Qiu.
\newblock Binary latent diffusion.
\newblock In \emph{Proceedings of the IEEE/CVF conference on computer vision and pattern recognition}, pages 22576--22585, 2023.

\bibitem[Weber et~al.()Weber, Yu, Yu, Deng, Shen, Cremers, and Chen]{webermaskbit}
Mark Weber, Lijun Yu, Qihang Yu, Xueqing Deng, Xiaohui Shen, Daniel Cremers, and Liang-Chieh Chen.
\newblock Maskbit: Embedding-free image generation via bit tokens.
\newblock \emph{Transactions on Machine Learning Research}.

\bibitem[Wu et~al.(2024{\natexlab{a}})Wu, Chen, Wu, Ma, Liu, Pan, Liu, Xie, Yu, Ruan, et~al.]{wu2024janus}
Chengyue Wu, Xiaokang Chen, Zhiyu Wu, Yiyang Ma, Xingchao Liu, Zizheng Pan, Wen Liu, Zhenda Xie, Xingkai Yu, Chong Ruan, et~al.
\newblock Janus: Decoupling visual encoding for unified multimodal understanding and generation.
\newblock \emph{arXiv preprint arXiv:2410.13848}, 2024{\natexlab{a}}.

\bibitem[Wu et~al.(2024{\natexlab{b}})Wu, Yin, Feng, He, Li, Hao, and Long]{wu2024ivideogpt}
Jialong Wu, Shaofeng Yin, Ningya Feng, Xu He, Dong Li, Jianye Hao, and Mingsheng Long.
\newblock ivideogpt: Interactive videogpts are scalable world models.
\newblock \emph{arXiv preprint arXiv:2405.15223}, 2024{\natexlab{b}}.

\bibitem[Wu et~al.(2024{\natexlab{c}})Wu, Zhang, Chen, Tang, Li, Fang, Zhu, Xie, Yin, Yi, et~al.]{wu2024vila}
Yecheng Wu, Zhuoyang Zhang, Junyu Chen, Haotian Tang, Dacheng Li, Yunhao Fang, Ligeng Zhu, Enze Xie, Hongxu Yin, Li Yi, et~al.
\newblock Vila-u: a unified foundation model integrating visual understanding and generation.
\newblock \emph{arXiv preprint arXiv:2409.04429}, 2024{\natexlab{c}}.

\bibitem[Xie et~al.(2024)Xie, Mao, Bai, Zhang, Wang, Lin, Gu, Chen, Yang, and Shou]{xie2024showo}
Jinheng Xie, Weijia Mao, Zechen Bai, David~Junhao Zhang, Weihao Wang, Kevin~Qinghong Lin, Yuchao Gu, Zhijie Chen, Zhenheng Yang, and Mike~Zheng Shou.
\newblock Show-o: One single transformer to unify multimodal understanding and generation.
\newblock \emph{arXiv preprint arXiv:2408.12528}, 2024.

\bibitem[Yu et~al.(2021)Yu, Li, Koh, Zhang, Pang, Qin, Ku, Xu, Baldridge, and Wu]{yu2021vector}
Jiahui Yu, Xin Li, Jing~Yu Koh, Han Zhang, Ruoming Pang, James Qin, Alexander Ku, Yuanzhong Xu, Jason Baldridge, and Yonghui Wu.
\newblock Vector-quantized image modeling with improved vqgan.
\newblock \emph{arXiv preprint arXiv:2110.04627}, 2021.

\bibitem[Yu et~al.()Yu, Lezama, Gundavarapu, Versari, Sohn, Minnen, Cheng, Gupta, Gu, Hauptmann, et~al.]{yulanguage}
Lijun Yu, Jose Lezama, Nitesh~Bharadwaj Gundavarapu, Luca Versari, Kihyuk Sohn, David Minnen, Yong Cheng, Agrim Gupta, Xiuye Gu, Alexander~G Hauptmann, et~al.
\newblock Language model beats diffusion-tokenizer is key to visual generation.
\newblock In \emph{The Twelfth International Conference on Learning Representations}.

\bibitem[Yu et~al.(2024{\natexlab{a}})Yu, He, Deng, Shen, and Chen]{yu2024randomized}
Qihang Yu, Ju He, Xueqing Deng, Xiaohui Shen, and Liang-Chieh Chen.
\newblock Randomized autoregressive visual generation.
\newblock \emph{arXiv preprint arXiv:2411.00776}, 2024{\natexlab{a}}.

\bibitem[Yu et~al.(2024{\natexlab{b}})Yu, Weber, Deng, Shen, Cremers, and Chen]{yu2024image}
Qihang Yu, Mark Weber, Xueqing Deng, Xiaohui Shen, Daniel Cremers, and Liang-Chieh Chen.
\newblock An image is worth 32 tokens for reconstruction and generation.
\newblock \emph{arXiv preprint arXiv:2406.07550}, 2024{\natexlab{b}}.

\bibitem[Zhang and Sennrich(2019)]{zhang2019root}
Biao Zhang and Rico Sennrich.
\newblock Root mean square layer normalization.
\newblock \emph{Advances in Neural Information Processing Systems}, 32, 2019.

\bibitem[Zhang et~al.(2018)Zhang, Isola, Efros, Shechtman, and Wang]{zhang2018unreasonable}
Richard Zhang, Phillip Isola, Alexei~A Efros, Eli Shechtman, and Oliver Wang.
\newblock The unreasonable effectiveness of deep features as a perceptual metric.
\newblock In \emph{Proceedings of the IEEE conference on computer vision and pattern recognition}, pages 586--595, 2018.

\bibitem[Zhu et~al.(2024)Zhu, Wei, Lu, and Chen]{zhu2024scaling}
Lei Zhu, Fangyun Wei, Yanye Lu, and Dong Chen.
\newblock Scaling the codebook size of vqgan to 100,000 with a utilization rate of 99\%.
\newblock \emph{arXiv preprint arXiv:2406.11837}, 2024.

\bibitem[Zhu et~al.()Zhu, Li, Zhang, Li, Xu, and Bing]{zhustabilize}
Yongxin Zhu, Bocheng Li, Hang Zhang, Xin Li, Linli Xu, and Lidong Bing.
\newblock Stabilize the latent space for image autoregressive modeling: A unified perspective.
\newblock In \emph{The Thirty-eighth Annual Conference on Neural Information Processing Systems}.

\end{thebibliography}
}


\end{document}